%% file: arxiv.tex
\documentclass[conference]{IEEEtran}
\IEEEoverridecommandlockouts
\usepackage{amsmath,amssymb,amsfonts}
\usepackage{algorithm}
\usepackage{algpseudocode}
\usepackage{graphicx}
\usepackage{textcomp}
\usepackage{xcolor}
\usepackage{multicol}
\usepackage{multirow}
\newsavebox{\tablebox}
\usepackage{ulem}
\usepackage{makecell}
\usepackage[misc,geometry]{ifsym} 
\usepackage{subfigure}
\usepackage{todonotes}
\setuptodonotes{inline}
\usepackage{amsmath}
\usepackage{tablefootnote}

\algnewcommand{\LineComment}[1]{\State \(\triangleright\) #1}

\newcommand{\Comments}[1]{}

\usepackage{mathtools}

\newcommand{\sig}[1]{{\small\textsf{{#1}}}}
\newcommand{\defas}{\coloneqq}

\IEEEoverridecommandlockouts                     

\PassOptionsToPackage{hyphens}{url}\usepackage{hyperref}

\algrenewcommand\algorithmicindent{0.75em}
\algrenewcommand\algorithmicthen{}
\algrenewcommand\algorithmicdo{}

\begin{document}
%

\title{Butterfly Effect Attack: Tiny and Seemingly Unrelated Perturbations for Object Detection}

\author{\IEEEauthorblockN{Nguyen Anh Vu Doan, Arda Yüksel, Chih-Hong Cheng\thanks{All authors have contributed equally, while the author ordering was computed randomly.  This work is supported by the Bavarian Ministry for Economic Affairs, Regional Development and Energy as part of a project to support the thematic development of the Fraunhofer IKS.}}	\IEEEauthorblockA{Fraunhofer IKS, Munich, Germany \\
 \{anhvu.doan, arda.yueksel, chih-hong.cheng\}@iks.fraunhofer.de
}	

\vspace{-10mm}
}

\maketitle

\begin{abstract}

This work aims to explore and identify tiny and seemingly unrelated perturbations of images in object detection that will lead to performance degradation. While tininess can naturally be defined using $L_p$ norms, we characterize the degree of ``unrelatedness'' of an object by the pixel distance between the occurred perturbation and the object. 
Triggering errors in prediction while satisfying two objectives  can be formulated as a multi-objective optimization problem where we utilize genetic algorithms to guide the search. The result successfully demonstrates that (invisible) perturbations on the right part of the image can drastically change the outcome of object detection on the left. An extensive evaluation reaffirms our conjecture that transformer-based object detection networks are more susceptible to butterfly effects in comparison to single-stage object detection networks such as YOLOv5.

\end{abstract}

\section{Introduction}

\begin{figure}[t]
	\centering

	\subfigure[The original image and the resulting prediction using YOLOv5]{
		\begin{minipage}[t]{\linewidth} 
\includegraphics[width=\linewidth]{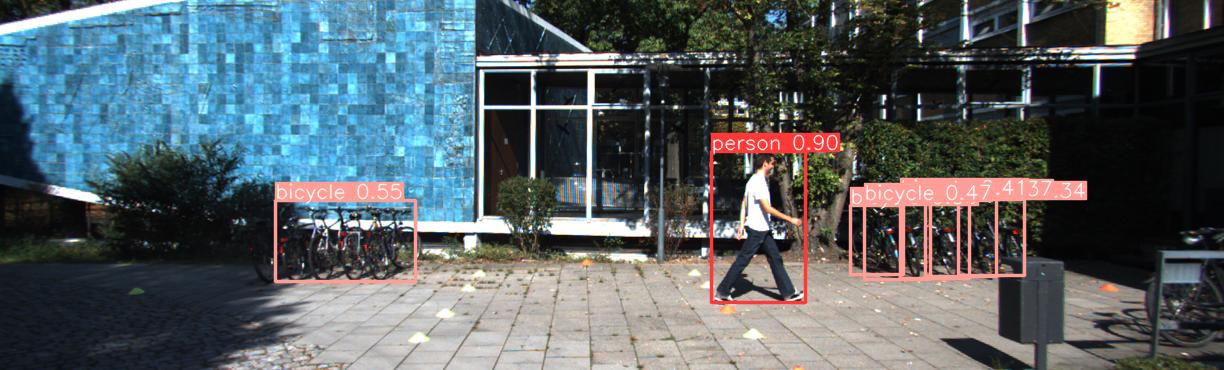}\label{fig:intro.original}
\end{minipage}
}\\%

\vspace{-2mm}

\subfigure[The perturbed image by adding noise only at the left part of the image]{
		\begin{minipage}[t]{\linewidth} 
\includegraphics[width=\linewidth]{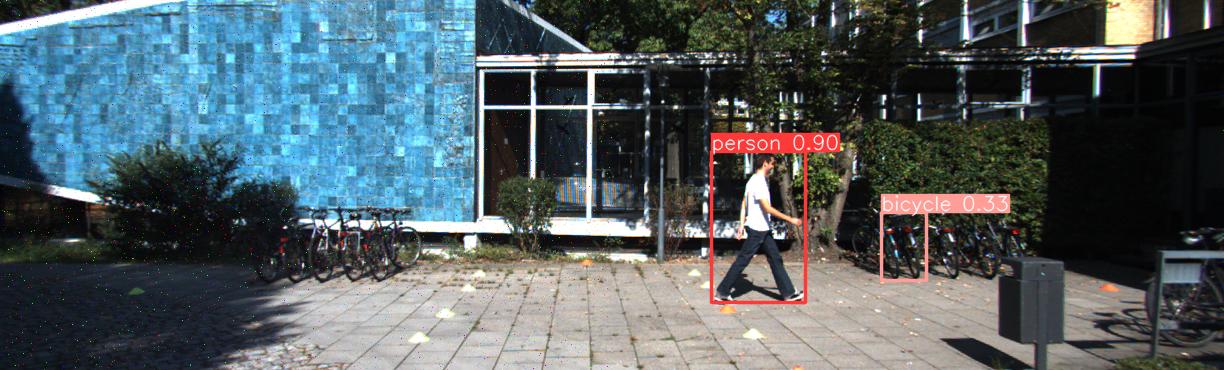}\label{fig:intro.after}
\end{minipage}
}

	\caption{Adding noises \textbf{only} on the left-hand side of the image  also leads to errors (missed bicycles) in the right-hand side.}
	\label{fig:introduction}
\end{figure}

Deep neural networks (DNNs) have been widely used in vision-based perception systems and are integral to realizing autonomous driving functions. Amongst various safety-related challenges such as data completeness or uncertainty quantification, we address the problem of robustness of DNNs. Concretely, we study object detection and consider perturbing an image to achieve a ``butterfly effect". By butterfly effect, we are not only addressing tiny perturbations but also interested in setting the perturbation on \emph{seemingly unrelated} locations of an image. The concept is illustrated using a real example in Figure~\ref{fig:introduction}. The perturbation changes the image (taken from the KITTI data set~\cite{geiger2012we}) from Figure~\ref{fig:intro.original} to Figure~\ref{fig:intro.after} by adding a slight perturbation on the left part of the image and by keeping the right part completely untouched. Still, the prediction generated by YOLOv5\footnote{The model (v6.1) is taken from the following website: \url{https://github.com/ultralytics/yolov5}} also leads to inconsistent results on the right-hand side. The consequence of successfully enabling such a type of perturbation implies that training by randomly adding noise over the complete image is insufficient for achieving robustness. Practically, it also hints that an attack on the moving vehicle in the front may be achieved by adding physical perturbation stickers on static objects on the side of the road. 

To find such a perturbation, we utilize a \emph{multi-objective optimization} approach based on the Non-dominated Sorting Genetic Algorithm (NSGA-II)~\cite{deb2002fast}. The goal of the multi-objective search is to simultaneously: 1) minimize the perturbation of the image; 2) maximize the performance degradation while considering multiple types of errors in both classification, bounding box size and location; 3) maximize the distance between the perturbation and the location of the detected object (or optionally, any object of the same class). 
Objective 3) is our formal definition of ``unrelated" which is included in the search process. We then interpret the results obtained with NSGA-II with the feature heatmap of the detection and integrate the comparison information into the NSGA-II exploration model. This approach allows \emph{interpretable} test case generation in that the translation from the specification to the objective is more direct. We demonstrate that the formulation allows easily reformulating the perturbation problem considering spatial redundancy (e.g., attacking an ensemble of networks~\cite{strauss2017ensemble}) as well as finding perturbation as filters to be continuously effective across multiple image frames.

While the internal connectivity of neural networks is the natural source of propagating small perturbations across regions, with the new test case generation technique, we aim to understand the impact by considering commonly seen architectural patterns for DNNs: Can transformer-based object detectors (e.g., DETR~\cite{carion2020end}), despite their high performance, be overall more resilient to such a butterfly effect compared to standard convolutional neural networks (e.g., YOLO~\cite{redmon2016you} and extensions)? An extensive evaluation reaffirms our conjecture that transformer-based object detection networks can be more susceptible to butterfly effects, potentially due to the attention mechanisms connecting two arbitrary regions in an image.

The rest of the paper is structured as follows: After reviewing existing results in Section~\ref{sec.related},  Section~\ref{sec.multi.objective} presents the key result, i.e., the encoding of the standard butterfly effect for object detection into  multiple objectives. Subsequently,  Section~\ref{sec.ga} considers how to use the genetic algorithms for our purpose, as well as summarizing extensions for designing perturbations against redundancy and temporal aspects. Finally, we summarize our experiment in Section~\ref{sec.evaluation} and conclude in Section~\ref{sec.concluding.remarks}.

\vspace{-2mm}

\section{Related Work}\label{sec.related}

Considering the whole spectrum of AI testing is beyond the scope of this paper. There exist foundational methods such as replay of previously collected challenging inputs~\cite{zendel2018wilddash}, adding random noises such as Gaussian or salt-and-pepper noises~\cite{hendrycks2019benchmarking,cheng2019nn}, or applying metamorphic relations (necessary conditions for output correctness) in testing~\cite{zhang2018deeproad,wang2019metamorphic}. The taxonomy of adversarial perturbation contains multiple dimensions, including attacker knowledge (white box or black box), the attack strategy (optimization or sensitivity analysis), or the attack goals (targeted or untargeted). 
Starting with initial concepts such as FGSM~\cite{szegedy2013intriguing} and the iterative version~\cite{kurakin2016adversarial}, Carlini \& Wagner attacks~\cite{carlini2017towards}, PGD~\cite{madry2017towards}, DeepFool~\cite{moosavi2016deepfool}, the problem of generating adversarial perturbation has been under active research. We refer readers to recent survey papers~\cite{huang2020survey,serban2020adversarial,xu2020adversarial} for references. Overall, our proposed method is a black box optimization-based approach. However, due to our encoding into the multi-objective optimization problem, we also can include feature-level distance as an additional optimization objective, thereby extending the approach to be a grey-box method. Other black box methods include direct adversarial image reuse from another model~\cite{papernot2016transferability}, transfer of an attack from separately learned ensembles~\cite{liu2016delving} or approximating gradients with finite difference methods~\cite{bhagoji2017exploring}. To this end, the closest to our work is GenAttack~\cite{alzantot2019genattack}, where the authors also applied a genetic algorithm as the black box perturbation framework. Our work differs from theirs in two key aspects: (1) Our focus is on object detection, while GenAttack is applied to classification. The nature of detecting multiple objects significantly increases the problem complexity, as there is a need to simultaneously cope with different error types. (2) We use a multi-objective optimization approach to cover diversified objectives such as the degree of ``unrelated" (which is a new dimension in object detection tasks), the amount of perturbation, and the change of prediction class within object detection. In contrast, GenAttack uses a single-objective optimization approach with the sole aim of changing the prediction class; controlling the amount of perturbation is set as an adaptive hyper-parameter that is not optimized explicitly.

\section{Butterfly Effects as Objectives}\label{sec.multi.objective}

\input{attack.tex}

\section{Realizing Butterfly Effect Attack with NSGA-II}\label{sec.ga}

\input{ga.tex}

\vspace{1mm}

\section{Evaluation}\label{sec.evaluation}

\input{evaluation.tex}

\section{Concluding Remarks}\label{sec.concluding.remarks}

In this work, we have investigated how a multi-objective optimization-based approach, with a customized NSGA-II algorithm, can be used to uncover performance degradation of object detectors while minimizing the intensity of perturbations on images. We defined objective functions to explore Pareto solutions between minimum image perturbation and maximum performance degradation while maximizing the distance between the image modifications and the objects. Experiments with the KITTI dataset were carried out on the YOLO and DETR object detectors by applying perturbations on the right side of images, and we analyzed the impact on the left side for single and ensemble of detector models (same perturbation on several YOLO/DETR models). While YOLO seems more robust than DETR in general, we observed that the object performance detection in both cases could be affected in different ways, even if the detected object itself has not been perturbed: 1) change in the bounding box and eventually a lower softmax score; 2) TP to FN; 3) TN to FP; 4) FN to TP; 5) FP to TN. This shows that butterfly effects unrelated to where an object is located can affect, positively or negatively, its detection.

For future work, we plan to refine further our mutation operation such that the initial mutation choices directly create human unrecognizable perturbation. Another dimension is considering how such an attack can be made physically available with optimization over new objectives such as different angles. Yet another direction is to consider how to utilize the developed technique to design new neural network architectures for robust object detection. Finally, we also plan to apply a similar paradigm to test motion planning modules.

\bibliographystyle{IEEEtran}

\end{document}

%% file: attack.tex
\subsection{Foundation}

We present an abstract definition for object detectors ignoring the internal connections of neurons as well as the associated pre- and post-processing mechanisms. An object detector is a function $f: \mathbb{R}^{L\times W\times 3} \rightarrow B^n$ that takes an input image in RGB of size $L\times W$ and generates a list of~$n$ bounding box predictions, with each prediction $B \defas (cl, x, y, l, w)$ outputting a bounding box of class~$cl \in \{1, \ldots, C\} \cup \{\bot\}$, centered at $(x, y)$ in the image plane with length~$l$ and width~$w$. The class $``\bot"$ is reserved for representing that the prediction does not contain an object; we call a bounding box prediction where $cl \neq \bot$ a \emph{valid} bounding box. 
Finally, we assume that given an image $\sig{img} \in \mathbb{R}^{L\times W\times 3}$, the generated prediction $f(\sig{img})$ is correct, and our goal is to generate a perturbation $\delta \in \mathbb{R}^{L\times W\times 3}$, such that $f(\sig{img} + \delta)$ generates degraded performance such as disappearing objects, ghost objects or incorrect bounding box size. The challenge here is that each type of degraded performance seems different. We have formulated an objective for degraded performance, which is detailed in the next section.

\subsection{Characterizing Butterfly Effects Using Three Objective Functions}

To create an attack that leads to butterfly effects, we aim to generate perturbations to simultaneously optimize three types of objectives.

\vspace{1mm}
\paragraph{Small perturbation} Generate a perturbation that is small in its quantity, thereby making it hard for a human to differentiate between the original image and the perturbed one. One can use different types of norms such as $L_1$, $L_2$ or $L_{\infty}$ to characterize the amount of perturbation being added. In this paper, we apply the $L_2$ norm, leading to the following objective function $\sig{obj}_{intensity}(\delta) \coloneqq \lVert \delta \rVert_2$.

\vspace{1mm}
\paragraph{Performance degradation} The second objective aims to generate prediction (for the perturbed image) being different from the original prediction (for the original image). As stated in the earlier section, the difference can arise from the change of output class, the size of the bounding box, and the location of the bounding box. Algorithm~\ref{algo.obj.performance.degradation} details how we design the objective function $\sig{obj}_{degrad}(\sig{img}, \delta, f)$ on characterizing the performance degradation when perturbing an image $\sig{img}$ with $\delta$. The variable~$A$ accumulates the sum of area overlap between the new and the old prediction. Starting at line~2, it considers every valid ($cl \neq \bot$) bounding box prediction~$B$ from the original prediction, finds the bounding box in the new prediction of the same type that has the largest area overlap (via computing the standard intersection-over-union metric $\sig{IoU}$ of two boxes\footnote{The intersection-over-union metric, also known as the Jaccard index, computes the ratio of the overlap and union areas between two bounding boxes. The calculated value is always between~$0$ and~$1$.} to variable~$AO$ (lines~3 to~8). Once when the largest area overlap is found, it is added to~$A$ to increase the sum (line~9). Finally, divide $A$ by the sum of valid ($cl \neq \bot$) bounding boxes (which equals the largest sum of the computed IoU metric) and use the computed value as the objective (line~11).

\begin{algorithm}[t]
\caption{Computing the objective $\sig{obj}_{degrad}$ }\label{algo.obj.performance.degradation}
    \begin{algorithmic}[1]
        \Require object detector $f$, input image $\sig{img}$, perturbation $\delta$
        \Ensure $\sig{obj}_{degrad}(\sig{img}, \delta, f)$
        \State let $A \gets 0$ 
       \ForAll{$B = (cl, x, y, l, w)  \in f(\sig{in})$ where $cl \neq \bot$}
       \State let $AO \gets 0$
        \ForAll{$B' = (cl', x', y', l', w')  \in f(\sig{in} + \delta)$}
        \If{$cl = cl'$ }
        \State $AO \gets \max (AO, \sig{IoU}(B, B'))$
        \EndIf
        \EndFor
        \State $A \gets A + AO$
        \EndFor
    \State \textbf{return} $\frac{A}{|\{B \;\defas (cl, l_x, l_y, l, w) \;|\; B \in f(\sig{in}) \; \wedge \; cl\neq \bot \}|}$
    \end{algorithmic} 
\end{algorithm}

Overall, an effective perturbation tends to \emph{lower} the objective $\sig{obj}_{degrad}$. To assist understanding the intuition behind, consider the simple case where the original prediction has only one valid bounding box. 

\begin{itemize}
    \item If the perturbed input does not lead to any change, then the computed objective equals~$1$.
    \item If the perturbed input leads to the bounding box changing its class to either~$\bot$ or to other class, then the ``if" statement at line~5 does not hold. As a consequence, $AO$ will not be updated at line~6, implying that~$AO$ remains to be~$0$. Therefore, the computed objective equals~$0$.
    \item If the perturbed input leads to the change of the size or the center position, then the computed IoU value is smaller than~$1$, and so is the computed objective.
\end{itemize}

\begin{algorithm}[t]
\caption{Computing the objective $\sig{obj}_{dist}$ }\label{algo.obj.distance.enlargement}
    \begin{algorithmic}[1]
        \Require object detector $f$, input image $\sig{img}$, perturbation $\delta$, buffer size $\epsilon$ to be used in surrounding the bounding box
        \Ensure $\sig{obj}_{dist}(\sig{img}, \delta, f)$
        \vspace{1mm}
       \State let $D \gets 0^{L\times W}$ \Comment{Initialize matrix $D$ with $0$s}
       \ForAll{$i \in \{1, \ldots, L\} \cap \mathbb{N}, j \in \{1, \ldots, W\} \cap \mathbb{N}$}
       \State $D[i,j] \gets \sqrt{L^2 + W^2}$ \Comment{Set to largest value}
       \ForAll{$B \defas (cl, x, y, l, w) \in f(\sig{img})$ where $cl \neq \bot$}
       \State $D[i,j] \gets \min (D[i,j], \sqrt{(x-i)^2 + (y-j)^2})$
       \EndFor        
        \EndFor

        \State $\sig{neg.avg} \gets (-1) \cdot \frac{\sum_{i,j} D[i,j]}{L\cdot W}$
        
        \ForAll{$i \in \{1, \ldots, L\} \cap \mathbb{N}, j \in \{1, \ldots, W\} \cap \mathbb{N}$}
       \ForAll{$B \defas (cl, x, y, l, w) \in f(\sig{img})$ where $cl \neq \bot$}
       \LineComment{If $(i,j)$ is inside a valid prediction box}
        \If{$i \in [x - \frac{l}{2} - \epsilon, x + \frac{l}{2} + \epsilon]$ and $j \in [y - \frac{w}{2} - \epsilon, y + \frac{w}{2} + \epsilon]$ }
        \State $D[i,j] \gets - \sig{neg.avg}$\Comment{Set to be negative average}
        \EndIf
       \EndFor
        \EndFor      
    
      \State let $\delta^{max}_{abs} \gets 0^{L\times W}$   
        \ForAll{$i \in \{1, \ldots, L\} \cap \mathbb{N}, j \in \{1, \ldots, W\} \cap \mathbb{N}$}
        \LineComment{Update $\delta^{max}_{abs}[i,j]$ w/ largest perturbation in RGB}
        \State $\delta^{max}_{abs}[i,j]  \gets \max(|\delta[i,j,1]|, |\delta[i,j,2]|, |\delta[i,j,3]|)$
        \State $D[i,j] \gets \delta^{max}_{abs}[i,j] \cdot D[i,j]$
       \EndFor
   \State $\sig{unpertubed.pixel.count} \gets \sum_{(i,j),  \delta^{max}_{abs}[i,j] \neq 0} 1$
    \State \textbf{return} $\frac{\sum_{i,j} D[i,j]}{\sig{unpertubed.pixel.count}}$
    \end{algorithmic} 
\end{algorithm}

\vspace{1mm}
\paragraph{Degree of unrelated perturbation} The final objective is to favor perturbations far from any valid bounding boxes in the original prediction. For example, suppose the bounding box is located at the center of the image. In this situation, we favor a perturbation that changes at the image's border rather than a perturbation that changes the center area. Algorithm~\ref{algo.obj.distance.enlargement} details how we design the objective function $\sig{obj}_{dist}(\sig{img}, \delta, f)$. Overall, an effective perturbation tends to increase the objective $\sig{obj}_{dist}$. This contrasts with the previously stated two objectives, where effective perturbation aims to decrease the objective in the previous cases.

Initially (lines~1 to~7), Algorithm~\ref{algo.obj.distance.enlargement} computes a matrix~$D$ which characterizes the minimum distance between any pixel $[i,j]$ to the center position of all valid bounding boxes. As our goal is to avoid perturbation within the bounding box, from lines~8 to~16, the algorithm scans through each pixel (line~9) and sets the value to be negative (reflected by the value computed at line~8) if the pixel is inside the box. The value~$\epsilon$ further adds a buffer and discourages the use of any perturbation surrounding the bounding box.  

Subsequently, the algorithm weighs the distance~$D[i,j]$ with the intensity of perturbation at pixel located at $[i,j]$. As we have RGB channels, line~20 computes the largest perturbation at pixel $[i,j]$, and use the computed value to weigh~$D[i,j]$ (line~21).  

Finally, the algorithm returns the computed objective (line~24) by using the sum of the weighted distance divided by the total number of unperturbed pixels (computed at line~23). Dividing the sum by the total number of unperturbed pixels turns out to be crucial in designing the objective. The intuition behind is as follows: it is possible to achieve the same weighted sum for two very different cases, namely

\begin{itemize}
    \item the case of having many tiny perturbations being nearby the object, and
    \item the case of having a relatively large perturbation on a few pixels being distant from any object. 
    
\end{itemize} 

The division by the total number of unperturbed pixels discourages the first scenario.

%% file: ga.tex
\subsection{Multi-Objective Optimization}

To perform the search, we apply the NSGA-II algorithm.
NSGA-II extends a classical genetic algorithm with two key concepts to enable multi-objective optimization: the \emph{Pareto rank} and the \emph{crowding distance} which allow a ``Pareto sorting" for the selection process.

\begin{itemize}
    \item The Pareto rank can be defined as follows. From a given pool of solutions, the Pareto optimal ones are of rank 1. For the higher ranks the following process is repeated iteratively: to find the solutions of rank $i \geq 2$, the solutions of rank $i-1$ are removed and the remaining Pareto solutions from this subset are of rank~$i$.
    
    \item The crowding distance is a measure of how close a point is to its neighbours, and thus reflects the density of solutions surrounding a particular point in the population. It is computed by taking the average distance of the two points on either side of this point along each of the objectives.
\end{itemize}

From the Pareto rank and the crowding distance, NSGA-II defines a ``Pareto sorting" for the selection process (modifying the classical binary tournament) as follows: between two solutions with different Pareto ranks, the lower rank will be preferred; otherwise, if both solutions have the same Pareto rank, then the one located in a less-crowded region will be preferred. We found that NSGA-II algorithm from the literature~\cite{deb2002fast} fits very well for our purposes, as we wish to select diversified solutions, which is reflected by the criterion ``less crowded". We have the following implementation choices for NSGA-II:

\vspace{1mm}
\paragraph{Encoding} We use an explicit encoding of the filter mask (matrix of modifications for the RGB values of each pixel) to perturb the images.

\vspace{1mm}
\paragraph{Initial population} The initial population consists of~$101$ individuals generated by randomly creating filter masks. Filter masks are taken as the individuals for the genetic algorithm. Our filter masks consist of signed integer values in the range of~$[-255, 255] \cap \mathbb{Z}$. The shape of the masks are defined according to the width and height of the images in the sequence. $100$ of these filter masks are randomly initialized from Gaussian distribution and later upon these masks various noise types of digital image processing are applied. In addition to that, a zero mask is added to the initial population (to keep the original image).

\vspace{1mm}
\paragraph{Crossover} One-point crossover is applied with a probability $p_c$ on the pixel array. Offspring filters are generated by using randomly sampled pixel indexes. Those pixel values are swapped and two offspring images are returned. 

\vspace{1mm}
\paragraph{Mutation} Generally, mutations are applied on genes; for our implementation, we refer pixels as individual genes of the filter masks. Four different mutation operations are investigated:
\begin{enumerate}
    \item Change the values of a random pixel with their complement in the $[-255, 255] \cap \mathbb{Z}$ range (similar to a bit flip).
    \item Shuffle randomly selected pixels (similar to a swap operation).
    \item Assign random values within $[-255, 255] \cap \mathbb{Z}$ for randomly sampled pixels. 
    \item Perform horizontal and/or vertical inversion of pixels.
\end{enumerate}

Finally, for all of these mutations, the modified pixels are taken from a parametrizable window size $w$. 

\subsection{Extensions Towards Attacking Ensembles and Temporally Stable Predictions}\label{sub.sec.extension}

With the proposed approach using a filter mask to model the image degradation, it is a straightforward extension to perform perturbation on a set of object detectors and deep ensembles, or even perform perturbation that is effective across frames. For ensembles, the insight is that the filter realizing the perturbation can be optimized such that the perturbation is effective across multiple predictors. 

In the following, we detail our defined objectives for ensembles via an aggregation of objectives from each detector. We exploit the notation and use the superscript ``$^k$" to denote the $k$-th object detector and its corresponding objectives, and assume there is a total of~$K$ detectors that form the ensemble. 

\vspace{-1mm}

\begin{equation}
    \sig{obj}^{ensemble}_{intensity} (\delta) \defas \sig{obj}^{1}_{intensity}(\delta) = \ldots = \sig{obj}^{K}_{intensity}(\delta)
\end{equation}

\begin{equation}
    \sig{obj}^{ensemble}_{degrad} \defas \frac{\sum_{k = 1 \ldots K} \sig{obj}_{degrad} (\sig{img}, \delta, f^k)}{K}
\end{equation}

\begin{equation}
    \sig{obj}^{ensemble}_{dist} \defas \frac{\sum_{k = 1 \ldots K} \sig{obj}_{dist} (\sig{img}, \delta, f^k)}{K}
\end{equation}

For $\sig{obj}^{ensemble}_{intensity}$, as the same mask (perturbation) is applied to all detectors within the ensemble, the objective for the ensemble is the same as the objective for individual detectors. For the remaining two objectives, their values are computed by averaging the individual objectives from constituting detectors in the ensemble.

Finally, for attacking temporally stable predictions, the single mask implementing~$\delta$ simply needs to be effective not on multiple predictors but rather on a sequence of images. Due to space limits, we omit detailing the formulation.

%% file: evaluation.tex
\begin{table}[t]
	\caption{Experiment parametrization}
	\label{tab:parameters:models.images}
	\centering
	\begin{tabular}{m{4cm}|c}
		Configuration & Value \\
		\hline
		$\#$ models generated & 25 YOLOv5 and 25 DETR \\
		$\#$ images tested on each model & 16 \\
		$\#$ models used in ensemble & 16 \\
		\hline
	\end{tabular}
	
\end{table}

\begin{table}[t]
	\caption{Configuration for NSGA-II}
	\label{tab:parameters:nsga-II}
	\centering
	\begin{tabular}{l|c}
		Parameter & Value \\
		\hline
		Number of iterations & 100 \\
		Population size & 101 \\ 
		Crossover probability & $p_c = 0.5$ \\
		Mutation probability & $p_m = 0.45$ \\
		Mutation window size\tablefootnote{Whenever a mutation is performed, at most 1\% of the pixels are affected.} & $w = 1\%$ \\
		\hline
	\end{tabular}
\end{table}

\begin{figure}[t]
	\centering
	\vspace{-10mm}
	\includegraphics[width=\linewidth]{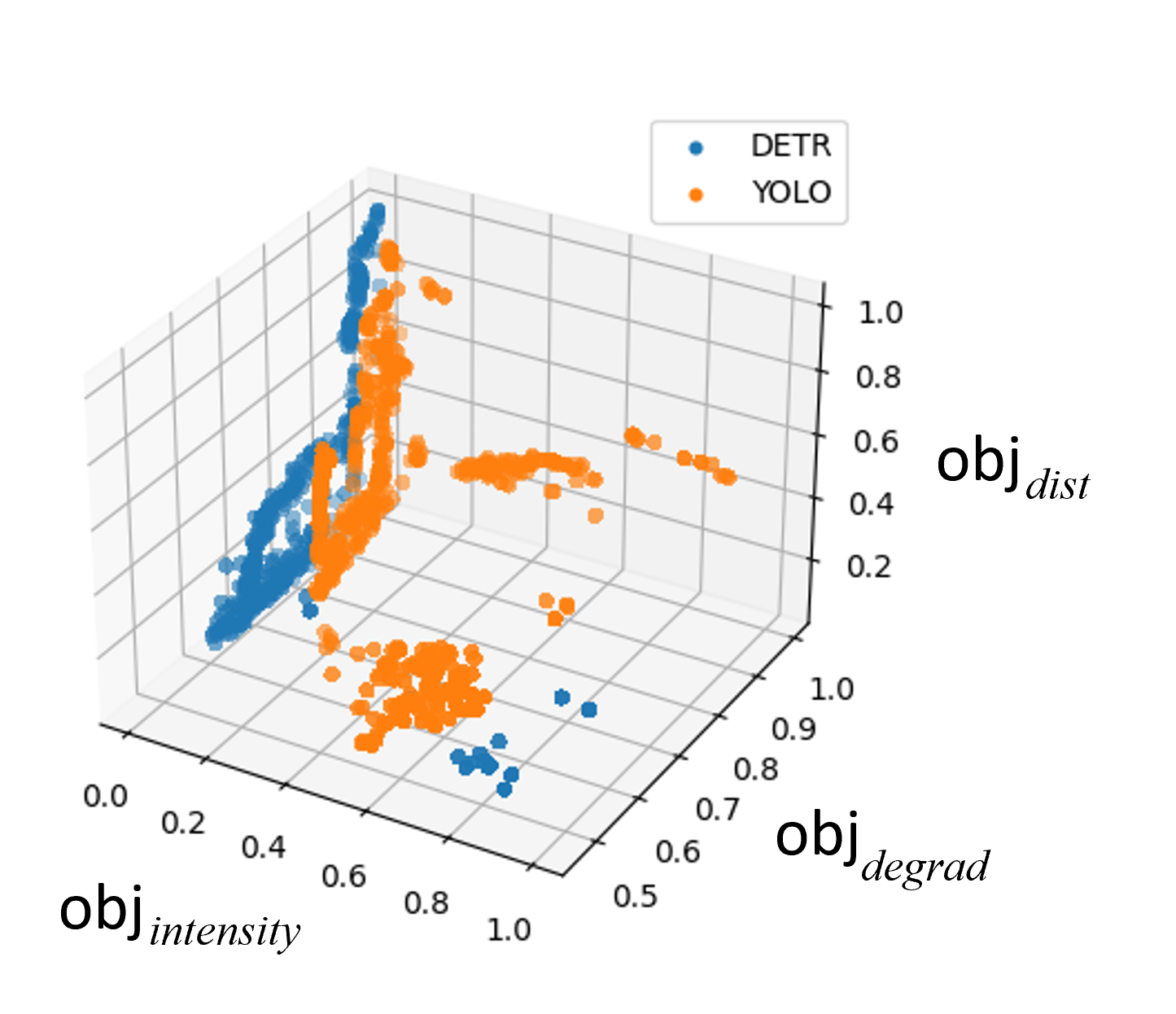}
	\vspace{-10mm}
	\caption{Comparing YOLO and DETR by visualizing three objectives at once}
	\label{fig:all_image_performance}
\end{figure}

\subsection{Overview and Experimental Setup}

We implemented our proposed method as a research prototype and tested the approach on two types of DNN-based object detectors under the KITTI dataset~\cite{geiger2012we}. The first type of detectors is the transformer (self-attention) network, where we use the DETR~\cite{carion2020end} model in our evaluation. The second type of detectors is the single-stage object detector extending convolutional neural networks, where we use YOLO version~5 as the evaluation target. 

Apart from demonstrating the existence of butterfly effect attacks, we are also interested in understanding whether transformer-based or single-stage convolutional object detectors are more vulnerable to butterfly effect attacks. Intuitively, due to the application of self-attention that connects two arbitrary regions in the decision-making, transformers can be more vulnerable to such attacks. Therefore, we have trained multiple models for each architecture using different random seeds\footnote{For repeatability, DETR and YOLO models are trained with random seeds $s \in [1, 25]$.} with perturbation applied on multiple images. We aim to derive the generic behavioral pattern between attention-based and single-stage detectors. Following the method described in Section~\ref{sub.sec.extension}, we also built ensembles and applied butterfly effect attacks. Table~\ref{tab:parameters:models.images} summarizes the models trained, the number of images fed into the detector, and the number of models used in building an ensemble. 

For the perturbation utilizing the NSGA-II algorithm, the parametrization of the evolutionary process is defined in Table~\ref{tab:parameters:nsga-II}. To highlight the qualitative effect of butterfly effects, we add a restriction where the perturbations are only applied to the right-hand side of the images, where we observe the resulting change on the left-hand side. This is done by forcing filters to have zeros in the left half (only the right half of an image is perturbed).

To help understanding the figures, we recall the criterion regarding a successful butterfly attack. For $\sig{obj}_{intensity}(\delta)$, the smaller the value, the more invisible the perturbation. For $\sig{obj}_{degrad} (\sig{img}, \delta, f)$ the smaller the value, the more effective the resulting performance drop. However, for $\sig{obj}_{dist} (\sig{img}, \delta, f)$ the larger the value, the further the perturbation is located.

\subsection{Comparing DETR and YOLO}

The result of applying NSGA-II on diverse models and images are highlighted in Figure~\ref{fig:all_image_performance}, where we only demonstrate resulting perturbations reflecting the Pareto fronts\footnote{In other words, although for the GA  maintains a population of $101$ elements when perturbing an image under a given architecture, we only show the resulting $3$ perturbations reflecting the best of three objectives with each being the best for one objective.} The result suggests that for DETR, with a smaller amount of perturbation, one can generate larger performance degradation. Also, we can observe that it is indeed possible to derive small perturbations such that one observes reasonable performance drop (the computed $\sig{obj}_{degrad}$ value is around $0.6$)  while the perturbation is seemly unrelated (the computed $\sig{obj}_{dist}$ is around $0.5$).  Here we omit further details, but in our evaluation, we also found that the attack method is equally applicable on ensembles.

Qualitatively, we have observed the following impacts caused by the butterfly effect attack:

\begin{figure}[t]
	\centering
	\subfigure[Original prediction]{
		\begin{minipage}[t]{\linewidth} 
			\includegraphics[width=\linewidth]{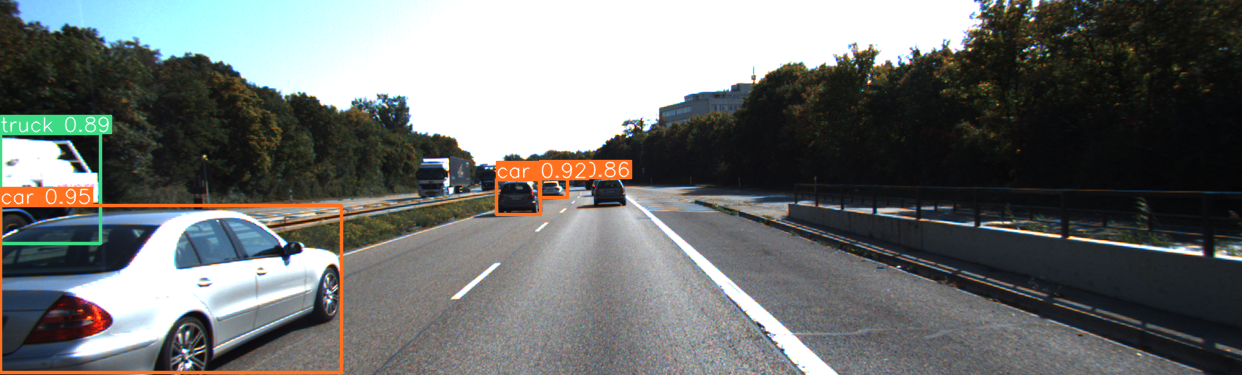}\label{fig:YOLO10_original}
		\end{minipage}
	}\\%
	
	\vspace{-2mm}
	
	\subfigure[Little performance degradation under strong noise]{
		\begin{minipage}[t]{\linewidth} 
			\includegraphics[width=\linewidth]{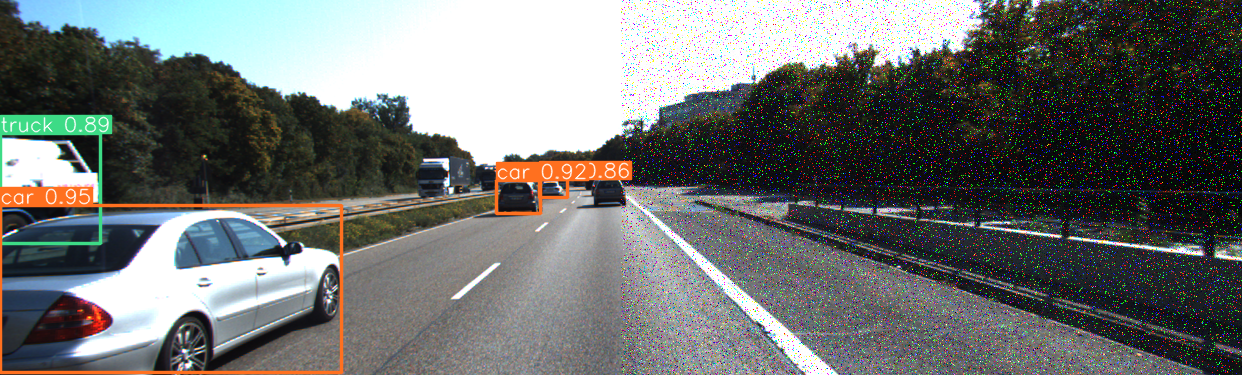}\label{fig:YOLO10_perform}
		\end{minipage}
	}

	\caption{Performance degradation with YOLO (image no. 10 of the KITTI data set); compared to Fig.~\ref{fig:DETR10_12perform} the high intensity of perturbation on the right does not lead to human-recognizable performance degradation on the left. }
	\label{fig:YOLO10}
\end{figure}

\begin{figure}
	\centering
	\subfigure[Original prediction]{
		\begin{minipage}[t]{\linewidth} 
			\includegraphics[width=\linewidth]{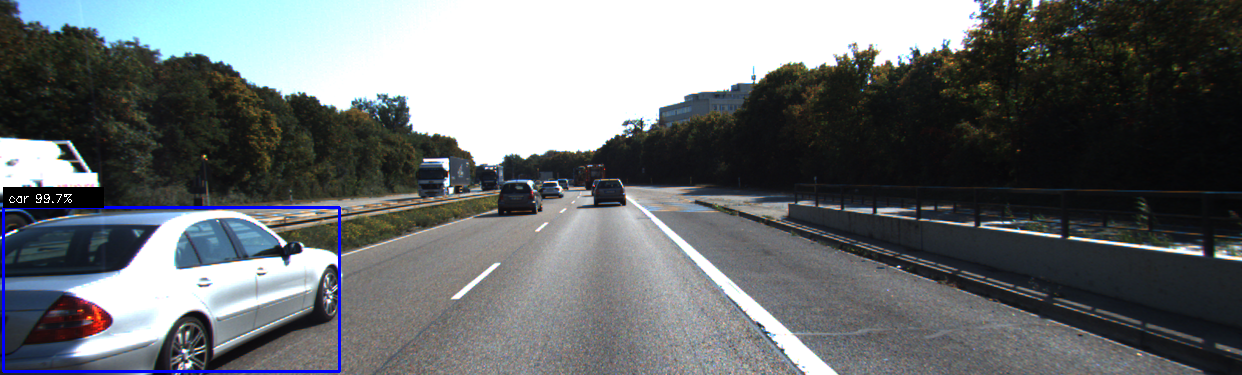}\label{fig:DETR10_12original}
		\end{minipage}
	}\\%
	
	\vspace{-2mm}
	
	\subfigure[Performance degradation by perturbing at the right-hand side ]{
		\begin{minipage}[t]{\linewidth} 
			\includegraphics[width=\linewidth]{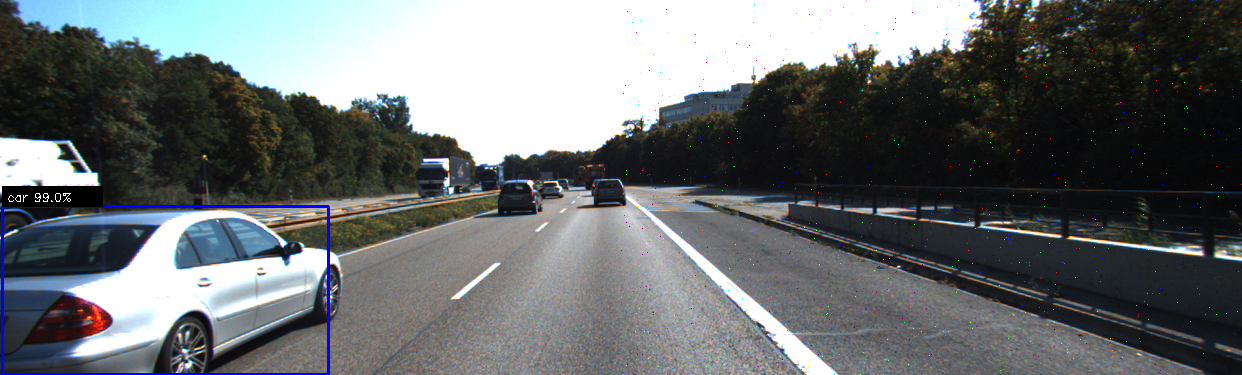}\label{fig:DETR10_12perform}
		\end{minipage}
	}
	
	\caption{Performance degradation with DETR (image no. 10 of the KITTI data set); compared to Fig.~\ref{fig:YOLO10_perform}, very small perturbation on the right already leads to  performance degradation (shrink of bounding box size) on the left. }
	\label{fig:DETR10_12}
\end{figure}

\begin{enumerate}
	\item The bounding box changes its size.
	\item True Positive (TP) becomes False Negative (FN), where a previously detected object is no longer detected. 
	\item True Negative (TN) becomes False Positive (FP), where the detector highlights the existence of a ghost object.
	\item False Negative (FN) becomes True Positive (TP), where a previously non-detected object becomes detected.
	\item False Positive (FP) becomes True Negative (TN), where a previous ghost object is no longer detected.
\end{enumerate}

\begin{figure}
	\centering
	\subfigure[Original prediction]{
		\begin{minipage}[t]{\linewidth} 
			\includegraphics[width=\linewidth]{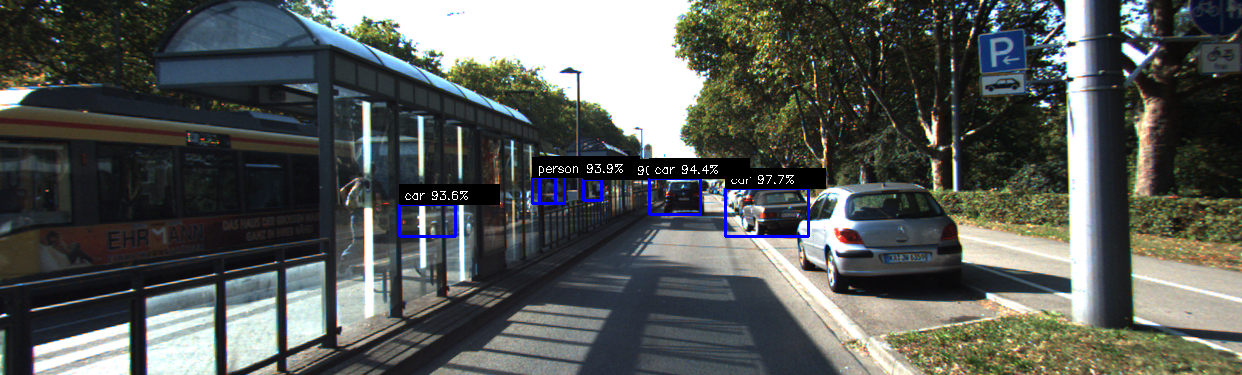}\label{fig:DETR7_22_original}
		\end{minipage}
	}\\%
	
	\vspace{-2mm}
	
	\subfigure[Resulting prediction by perturbing at the right-hand side ]{
		\begin{minipage}[t]{\linewidth} 
			\includegraphics[width=\linewidth]{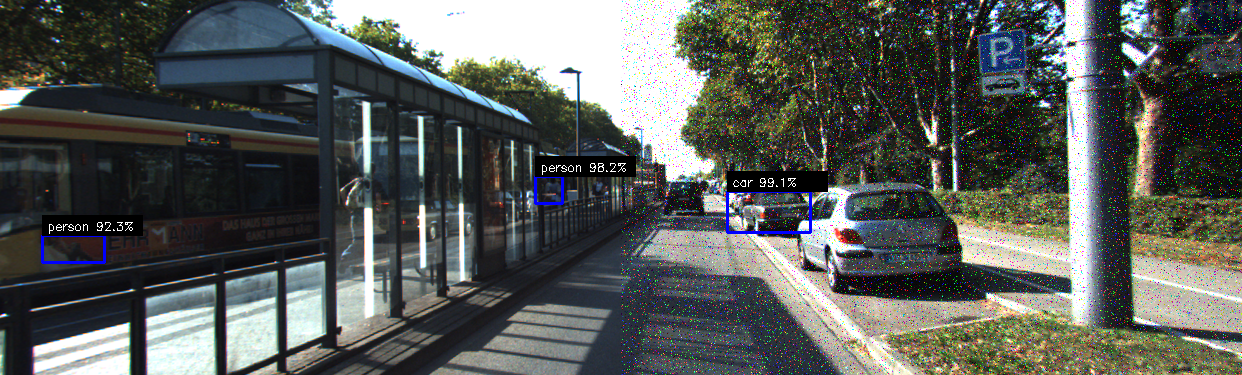}\label{fig:DETR7_22perform}
	\end{minipage}}\\%
	
	\caption{True negative becomes false positive: non-existing person object appears at the left of the image}
	\label{fig:DETR7_22}
\end{figure}

Figure~\ref{fig:YOLO10} and~\ref{fig:DETR10_12} show the required perturbation on the same image between YOLO and DETR. For YOLO (Figure~\ref{fig:YOLO10}), even when the perturbation intensity on the right is already human-recognizable, the resulting prediction remains the same. This is in contrast to the DETR detector (Figure~\ref{fig:DETR10_12}) where a small perturbation on the right already leads to the change of the bounding box of the left car. Another example in 
Figure~\ref{fig:DETR7_22} highlights the situation where the butterfly effect attack generates ghost objects (recall that the left hand side of the image is completely un-modified). Finally, the example in Figure~\ref{fig:introduction} demonstrates the situation of disappearing objects.

%% file: arxiv.bbl
\begin{thebibliography}{10}
	\providecommand{\url}[1]{#1}
	\csname url@rmstyle\endcsname
	\providecommand{\newblock}{\relax}
	\providecommand{\bibinfo}[2]{#2}
	\providecommand\BIBentrySTDinterwordspacing{\spaceskip=0pt\relax}
	\providecommand\BIBentryALTinterwordstretchfactor{4}
	\providecommand\BIBentryALTinterwordspacing{\spaceskip=\fontdimen2\font plus
		\BIBentryALTinterwordstretchfactor\fontdimen3\font minus
		\fontdimen4\font\relax}
	\providecommand\BIBforeignlanguage[2]{{%
			\expandafter\ifx\csname l@#1\endcsname\relax
			\typeout{** WARNING: IEEEtran.bst: No hyphenation pattern has been}%
			\typeout{** loaded for the language `#1'. Using the pattern for}%
			\typeout{** the default language instead.}%
			\else
			\language=\csname l@#1\endcsname
			\fi
			#2}}
	
	\bibitem{geiger2012we}
	A.~Geiger, P.~Lenz, and R.~Urtasun, ``Are we ready for autonomous driving? the
	{KITTI} vision benchmark suite,'' in \emph{CVPR}.\hskip 1em plus 0.5em minus
	0.4em\relax IEEE, 2012, pp. 3354--3361.
	
	\bibitem{deb2002fast}
	K.~Deb, A.~Pratap, S.~Agarwal, and T.~Meyarivan, ``{A fast and elitist
		multiobjective genetic algorithm: NSGA-II},'' \emph{IEEE transactions on
		evolutionary computation}, vol.~6, no.~2, pp. 182--197, 2002.
	
	\bibitem{strauss2017ensemble}
	T.~Strauss, M.~Hanselmann, A.~Junginger, and H.~Ulmer, ``Ensemble methods as a
	defense to adversarial perturbations against deep neural networks,''
	\emph{arXiv preprint arXiv:1709.03423}, 2017.
	
	\bibitem{carion2020end}
	N.~Carion, F.~Massa, G.~Synnaeve, N.~Usunier, A.~Kirillov, and S.~Zagoruyko,
	``End-to-end object detection with transformers,'' in \emph{ECCV}.\hskip 1em
	plus 0.5em minus 0.4em\relax Springer, 2020, pp. 213--229.
	
	\bibitem{redmon2016you}
	J.~Redmon, S.~Divvala, R.~Girshick, and A.~Farhadi, ``You only look once:
	Unified, real-time object detection,'' in \emph{CVPR}, 2016, pp. 779--788.
	
	\bibitem{zendel2018wilddash}
	O.~Zendel, K.~Honauer, M.~Murschitz, D.~Steininger, and G.~F. Dominguez,
	``Wilddash-creating hazard-aware benchmarks,'' in \emph{ECCV}, 2018, pp.
	402--416.
	
	\bibitem{hendrycks2019benchmarking}
	D.~Hendrycks and T.~Dietterich, ``Benchmarking neural network robustness to
	common corruptions and perturbations,'' \emph{arXiv preprint
		arXiv:1903.12261}, 2019.
	
	\bibitem{cheng2019nn}
	C.-H. Cheng, C.-H. Huang, and G.~N{\"u}hrenberg, ``nn-dependability-kit:
	Engineering neural networks for safety-critical autonomous driving systems,''
	in \emph{ICCAD}.\hskip 1em plus 0.5em minus 0.4em\relax IEEE, 2019, pp. 1--6.
	
	\bibitem{zhang2018deeproad}
	M.~Zhang, Y.~Zhang, L.~Zhang, C.~Liu, and S.~Khurshid, ``Deeproad: Gan-based
	metamorphic testing and input validation framework for autonomous driving
	systems,'' in \emph{ASE}.\hskip 1em plus 0.5em minus 0.4em\relax IEEE, 2018,
	pp. 132--142.
	
	\bibitem{wang2019metamorphic}
	S.~Wang and Z.~Su, ``Metamorphic testing for object detection systems,''
	\emph{arXiv preprint arXiv:1912.12162}, 2019.
	
	\bibitem{szegedy2013intriguing}
	C.~Szegedy, W.~Zaremba, I.~Sutskever, J.~Bruna, D.~Erhan, I.~Goodfellow, and
	R.~Fergus, ``Intriguing properties of neural networks,'' \emph{arXiv preprint
		arXiv:1312.6199}, 2013.
	
	\bibitem{kurakin2016adversarial}
	A.~Kurakin, I.~Goodfellow, and S.~Bengio, ``Adversarial machine learning at
	scale,'' \emph{arXiv preprint arXiv:1611.01236}, 2016.
	
	\bibitem{carlini2017towards}
	N.~Carlini and D.~Wagner, ``Towards evaluating the robustness of neural
	networks,'' in \emph{S\&P}.\hskip 1em plus 0.5em minus 0.4em\relax Ieee,
	2017, pp. 39--57.
	
	\bibitem{madry2017towards}
	A.~Madry, A.~Makelov, L.~Schmidt, D.~Tsipras, and A.~Vladu, ``Towards deep
	learning models resistant to adversarial attacks,'' \emph{arXiv preprint
		arXiv:1706.06083}, 2017.
	
	\bibitem{moosavi2016deepfool}
	S.-M. Moosavi-Dezfooli, A.~Fawzi, and P.~Frossard, ``Deepfool: a simple and
	accurate method to fool deep neural networks,'' in \emph{CVPR}, 2016, pp.
	2574--2582.
	
	\bibitem{huang2020survey}
	X.~Huang, D.~Kroening, W.~Ruan, J.~Sharp, Y.~Sun, E.~Thamo, M.~Wu, and X.~Yi,
	``A survey of safety and trustworthiness of deep neural networks:
	Verification, testing, adversarial attack and defence, and
	interpretability,'' \emph{Computer Science Review}, vol.~37, p. 100270, 2020.
	
	\bibitem{serban2020adversarial}
	A.~Serban, E.~Poll, and J.~Visser, ``Adversarial examples on object
	recognition: A comprehensive survey,'' \emph{ACM Computing Surveys (CSUR)},
	vol.~53, no.~3, pp. 1--38, 2020.
	
	\bibitem{xu2020adversarial}
	H.~Xu, Y.~Ma, H.-C. Liu, D.~Deb, H.~Liu, J.-L. Tang, and A.~K. Jain,
	``Adversarial attacks and defenses in images, graphs and text: A review,''
	\emph{International Journal of Automation and Computing}, vol.~17, no.~2, pp.
	151--178, 2020.
	
	\bibitem{papernot2016transferability}
	N.~Papernot, P.~McDaniel, and I.~Goodfellow, ``Transferability in machine
	learning: from phenomena to black-box attacks using adversarial samples,''
	\emph{arXiv preprint arXiv:1605.07277}, 2016.
	
	\bibitem{liu2016delving}
	Y.~Liu, X.~Chen, C.~Liu, and D.~Song, ``Delving into transferable adversarial
	examples and black-box attacks,'' \emph{arXiv preprint arXiv:1611.02770},
	2016.
	
	\bibitem{bhagoji2017exploring}
	A.~N. Bhagoji, W.~He, B.~Li, and D.~Song, ``Exploring the space of black-box
	attacks on deep neural networks,'' \emph{arXiv preprint arXiv:1712.09491},
	2017.
	
	\bibitem{alzantot2019genattack}
	M.~Alzantot, Y.~Sharma, S.~Chakraborty, H.~Zhang, C.-J. Hsieh, and M.~B.
	Srivastava, ``Genattack: Practical black-box attacks with gradient-free
	optimization,'' in \emph{GECCO}, 2019, pp. 1111--1119.
	
\end{thebibliography}
